\documentclass[10pt,twocolumn,letterpaper]{article}

\usepackage{cvpr}              

\usepackage{graphicx}
\usepackage{amsmath}
\usepackage{amssymb}
\usepackage{booktabs}

\newcommand{\fssupptitle}{\fontsize{15pt}{\baselineskip}\selectfont}
\newcommand{\appendixhead}%
{\centering\textbf{\fssupptitle Supplementary Material}
\vspace{0.25in}}

\usepackage[utf8]{inputenc} 
\usepackage[T1]{fontenc}    
\usepackage{hyperref}       
\usepackage{url}            
\usepackage{booktabs}       
\usepackage{amsfonts}       
\usepackage{nicefrac}       
\usepackage{microtype}      
\usepackage{xcolor}         

\usepackage{times}
\usepackage{epsfig}
\usepackage{amsmath}
\usepackage{amssymb}
\usepackage{bold-extra}

\newcommand{\rz}[1]{{\color{black}#1}}
\newcommand{\rizi}[1]{{\color{black}#1}}

\usepackage[capitalize]{cleveref}
\crefname{section}{Sec.}{Secs.}
\Crefname{section}{Section}{Sections}
\Crefname{table}{Table}{Tables}
\crefname{table}{Tab.}{Tabs.}

\def\cvprPaperID{6145} 
\def\confName{CVPR}
\def\confYear{2023}

\begin{document}

\title{ARO-Net: Learning Neural Fields from Anchored Radial Observations}
\title{ARO-Net: Learning \rz{Implicit} Fields from Anchored Radial Observations}

\author{Yizhi Wang$^{1,2}$\thanks{Equal contribution}, Zeyu Huang$^1$\footnotemark[1], Ariel Shamir$^3$, Hui Huang$^1$, Hao Zhang$^2$, Ruizhen Hu$^1$\thanks{Corresponding author. E-mail: ruizhen.hu@gmail.com}\\
$^1$Shenzhen University
$^2$Simon Fraser University
$^3$Reichman University}

\maketitle

\begin{abstract}

We introduce {\em anchored radial observations\/} (ARO), a novel shape encoding for learning \rz{implicit} field representation of \rz{3D} shapes that 
is {\em category-agnostic\/} and {\em generalizable\/} amid significant shape variations.
The main idea behind our work is to reason about shapes through partial observations from a set of 
viewpoints, called anchors.
We develop a general and unified shape representation by employing a fixed set of anchors, via Fibonacci sampling, and designing a coordinate-based 
deep neural network to predict the occupancy value of a query point in space.
Differently from prior neural implicit models that use global shape feature, our shape encoder operates on {\em contextual, query-specific\/} features. To predict point occupancy, locally 
observed shape information from the perspective of the anchors surrounding the input query point are encoded and aggregated through an attention module, before implicit decoding is performed.
We demonstrate the quality and generality of our network, coined ARO-Net, on surface reconstruction from sparse point clouds,
with tests on novel and unseen object categories, ``one-shape'' training, and comparisons to state-of-the-art neural and classical 
methods for reconstruction and tessellation.
Our code is available at: \href{https://github.com/yizhiwang96/ARO-Net}{https://github.com/yizhiwang96/ARO-Net}.
\end{abstract}

\section{Introduction}
\label{sec:intro}


Despite the substantial progress made in deep learning in recent years, transferability and generalizability issues still
persist due to domain shifts and the need to handle diverse, out-of-distribution test cases. For 3D shape representation
learning, a reoccurring challenge has been the inability of trained neural models to generalize to unseen object categories
and diverse shape structures, as these models often overfit to or ``memorize" the training data.

\begin{figure}
\centering
  \includegraphics[width=0.99\linewidth]{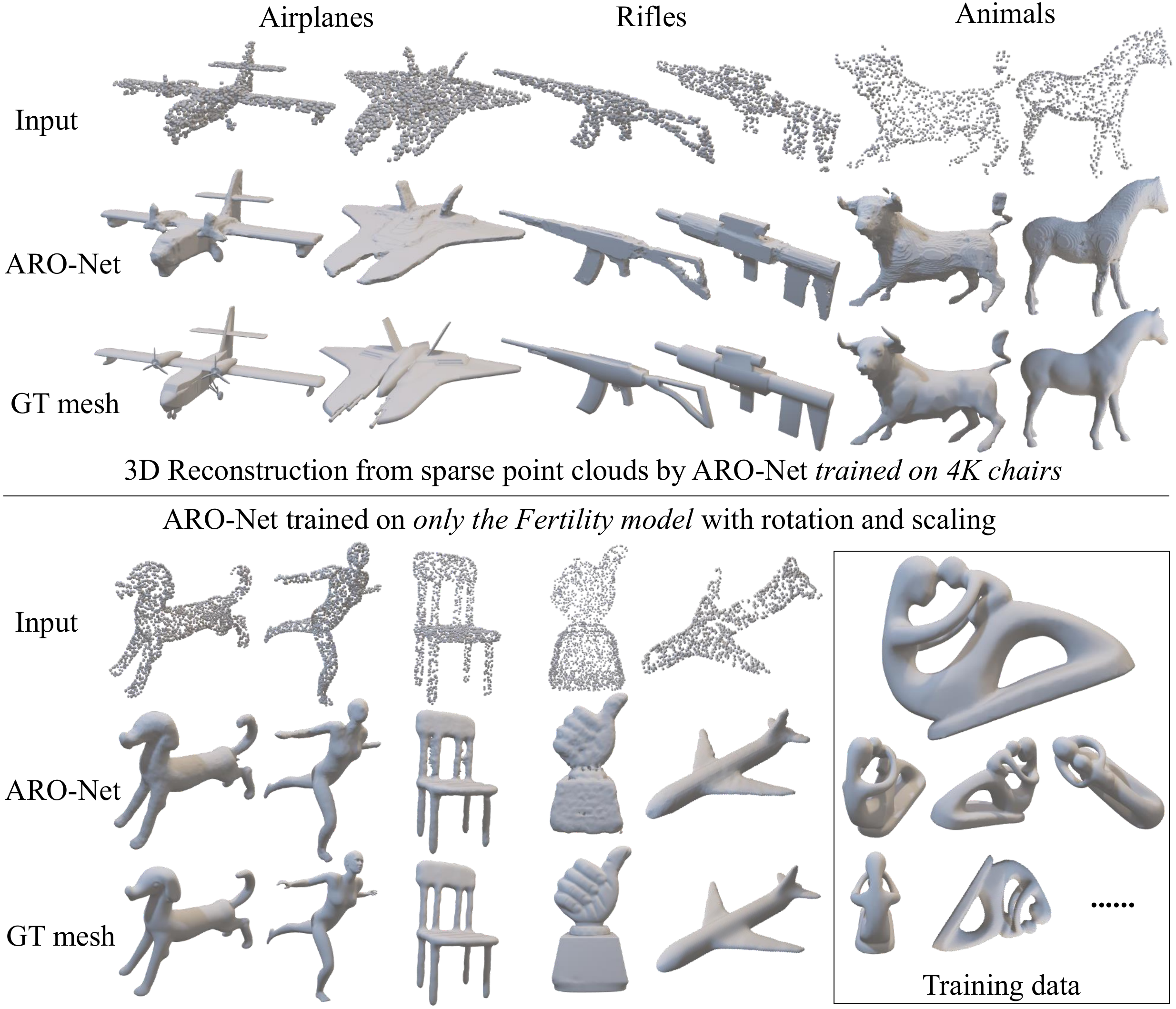}
  \caption{Neural 3D reconstruction using ARO-Net from sparse point clouds (1,024 or 2,048 points). Top: reconstruction results on airplanes, riffles, and animals when the network was trained only on chairs. Bottom: reconstruction of a variety of shapes when the network was train on numerous versions of one model - the Fertility. See comparison to other methods in Section~\ref{sec:results}.}
  \label{fig:teaser}
\end{figure}

In this paper, we introduce a novel shape encoding for learning an \rz{implicit} field~\cite{NF_survey} representation of \rz{3D} shapes
that is {\em category-agnostic\/} and {\em generalizable\/} amid significant shape variations.
In Figure~\ref{fig:teaser} (top), we show 3D reconstructions from sparse point clouds
obtained by our approach that is trained on chairs but tested on airplanes, riffles, and animals.
Our model can even reconstruct a variety of shapes when the training data consists of 
{\em only one\/} shape, augmented with rotation and scaling; see Figure~\ref{fig:teaser} (bottom).

The main idea behind our work is to reason about shapes from {\em partial observations\/} at a set of 
viewpoints, called {\em anchors\/}, and apply this reasoning to learn \rz{implicit} fields.
%
We develop a general and unified shape representation by designating a {\em fixed\/} set of
anchors and designing a coordinate-based neural network to predict the occupancy at a query point.
In contrast to classical neural implicit models such as IM-Net~\cite{IM-Net}, OccNet~\cite{mescheder2019occupancy}, and 
DeepSDF~\cite{park2019deepsdf}, which learn global shape features for occupancy/distance prediction, our novel
encoding scheme operates on {\em local\/} shape features obtained by viewing the query point from the anchors.

\begin{figure*}[!t]
    \centering
    \includegraphics[width=0.99\textwidth]{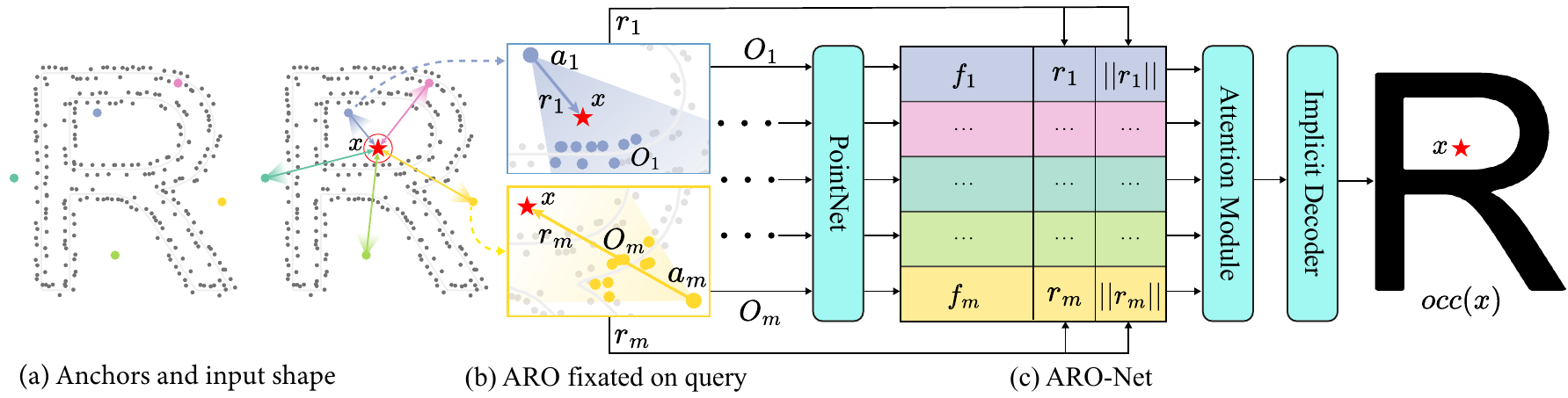}
\caption{2D illustration of ARO and ARO-Net architecture: (a) Input point cloud (in grey) and a set of $m$ fixed anchors (coloured dots). (b) Radial observation $O_i$ from each anchor $a_i$ toward the query point $x$ consists of closed points inside the cone apexed at $a_i$, with axis $r_i = x - a_i $. (c) Each radial observation $O_i$ is passed to a PointNet encoder to obtain an embedding feature $f_i$, which is concatenated with $r_i$ and its norm to form the query-specific ARO encoding of $x$ with respect to $a_i$. Finally, all the ARO features are decoded into the occupancy value $occ(x)$ though an attention module and several MLPs. For 3D reconstruction, the ARO features are computed for each query point, while the PointNet, attention module, and implicit decoder are fixed during inference --  their weights were determined during training.}
\label{fig:aro_net}
\end{figure*}

\begin{figure}
\centering
  \includegraphics[width=0.99\linewidth]{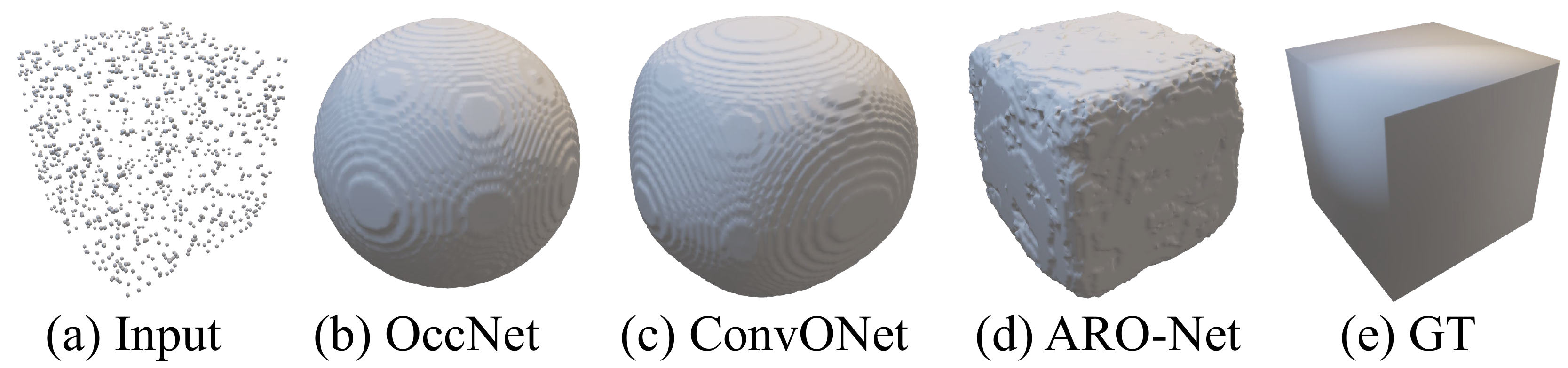}
  \caption{A somewhat extreme toy example comparing ARO-Net to prior occupancy prediction networks on 3D reconstruction from a sparse point cloud of a cube (a), with training on a single sphere. The results from OccNet~\cite{mescheder2019occupancy} and ConvONet)~\cite{Peng2020_CON} show more signs of overfitting to the training sphere than ARO-Net (d).}
  \label{fig:AROvsOCC}
\end{figure}

Specifically, for a given query point $x$, we collect locally observable shape information surrounding $x$, as well as directional 
and distance information toward $x$, from the perspective of 
the set of anchors, and encode such information using PointNet~\cite{PointNet}; see Figure~\ref{fig:aro_net}(b). The PointNet features 
are then aggregated through an attention module, whose output is fed to an implicit decoder to produce the occupancy value 
for $x$. We call our {\em query-specific\/} shape encoding {\em Anchored Radial Observations\/}, or {\em ARO\/}. The prediction 
network is coined ARO-Net, as illustrated in Figure~\ref{fig:aro_net}.

The advantages of ARO for learning \rz{implicit} fields are three-fold. First, shape inference from partial and local observations is not 
bound by barriers set by object categories or structural variations. Indeed, local shape features are more prevalent, and 
hence more generalizable, across categories than global ones~\cite{IM-Net,mescheder2019occupancy,park2019deepsdf}. 
Second, ARO is query-specific and the directional and distance information it includes is intimately tied to the occupancy
prediction at the query point, as explained in Section~\ref{sec:aro_rep}.
Last but not least, by aggregating observations from all the anchors, the resulting encoding is not purely local, like the voxel-level
encoding or latent codes designed to better capture geometric details across large-scale scenes~\cite{Jiang2020_LocalGrid,Peng2020_CON,Chabra2020_DeepLS}. 
ARO effectively captures more global and {\em contextual\/} query-specific shape features. 

In Figure~\ref{fig:AROvsOCC}, we demonstrate using a toy example the difference between ARO-Net and representative 
neural implicit models based on global (OccNet~\cite{mescheder2019occupancy}) and local grid shape encodings (convolutional occupancy
network or ConvONet~\cite{Peng2020_CON}), for the 3D reconstruction task. All three networks were trained on a single sphere shape, with a single 
anchor for ARO inside the sphere. When tested on reconstructing a cube, the results show that both OccNet and ConvONet exhibit more 
memorization of the training sphere either globally or locally, while the ARO-Net reconstruction is more faithful to the input without special fine-tuning.

In our current implementation of ARO-Net, we adopt Fibonacci sampling~\cite{keinert2015spherical} to obtain the fixed set of anchors. We 
demonstrate the quality and generalizability of our method on surface reconstruction from sparse point clouds, with testing on novel and 
unseen object categories, as well as ``one-shape'' training (see bottom of Figure~\ref{fig:teaser}). We report extensive quantitative and 
qualitative comparison results to state-of-the-art methods including both neural models for reconstruction~\cite{IM-Net,mescheder2019occupancy,Peng2020_CON,Points2Surf,BPS} and tessellation~\cite{NDC}, as well as classical schemes 
such as screen Poisson reconstruction~\cite{SPR}. Finally, we conduct ablation studies to evaluate our choices for
the number of anchors, the selection strategies, and the decoder architecture: MLP vs.~attention modules. 

\section{Related work}
\label{sec:related}

Learning neural fields is an emerging and one of the most intensely studied subjects in the vision and machine learning communities 
in recent years, while surface reconstruction is one of the most classical problems in computer vision and graphics. In this section, we 
shall only focus on the most closely related prior works. For a more comprehensive coverage on the relevant topics, we refer the readers 
to the surveys on surface reconstruction from point clouds~\cite{ReconSurvey} and neural fields~\cite{NF_survey}, as well as the the 
excellent \href{https://github.com/vsitzmann/awesome-implicit-representations}{summary on neural implicit representations} by Vincent Sitzmann.

\vspace{-8pt}

\paragraph{Surface reconstruction and local priors.}
%
Classical reconstruction schemes such as Screened Poisson (SPR)~\cite{SPR} often produce artifacts amid noise and 
under sampling. One remedy is to rely on {\em local shape priors\/}, which are reoccurring patch templates 
that can be matched to help reconstruction over sparse input.
Gal et al.~\cite{Gal_SGP2007} develop such an example-based surface reconstruction method, relying on a pre-built set
of enriched patches sampled from a database of 3D models that include surface normals and point feature classification 
(e.g., edge or corner) to facilitate reconstruction. These patches essentially serve as a training set.

Taking local templates to the extreme gives us classical tessellation methods such as Marching Cubes~\cite{MC} and 
Dual Countouring~\cite{DC}, whose ``neural'' versions have been developed recently as neural Marching Cubes~\cite{NMC} 
and neural Dual Contouring (NDC)~\cite{NDC}. In particular, the unsigned version of NDC, or UNDC,
can achieve state-of-the-art results on reconstruction from un-oriented point clouds with sufficient sampling. Indeed, 
what is common about all of these purely local reconstruction schemes is their ability to generalize 
across object categories, while disregarding any contextual shape information and global reconstruction priors.

\vspace{-8pt}

\paragraph{Neural implicit models.}
%
%
Since OccNet~\cite{mescheder2019occupancy}, IM-Net~\cite{IM-Net}, and DeepSDF~\cite{park2019deepsdf} concurrently 
introduced the coordinate-based implicit neural representation in 2019, research and development on neural \rz{implicit} fields~\cite{NF_survey} have 
flourished. In these early works, the shape encoder~\cite{mescheder2019occupancy,IM-Net} and latent 
code~\cite{park2019deepsdf} employed for occupancy/distance predictions only model global shape features. The
resulting learned representations were unable to reconstruct fine geometric details.

The next wave of developments has focused on conditioning implicit neural representations on local features stored in 
voxel~\cite{Chibane2020,Jiang2020_LocalGrid,Peng2020_CON,Chabra2020_DeepLS} or image grids~\cite{PIFu,DISN,D2IM-Net}
to more effectively recover geometric or topological details and to scale to scene reconstruction, 
or at the patch level~\cite{LDIF,PatchNets} to improve generalizability across object categories. 
%

Some of these locally conditioned neural implicit models essentially learn local shape priors~\cite{Gal_SGP2007} from training 
shapes, where the learned priors are parameterized in different ways, e.g., voxel-level latent codes in Deep Local 
Shape~\cite{Chabra2020_DeepLS}, embeddings of ShapeNet~\cite{ShapeNet} parts in local implicit grid representations~\cite{Jiang2020_LocalGrid}, 
structured Gaussians in Local Deep Implicit Functions (LDIF)~\cite{LDIF}, and canonical patches in PatchNets~\cite{PatchNets}.
Other works~\cite{Chibane2020,Peng2020_CON,Points2Surf}, like ARO-Net, also utilize query-specific shape features.

\vspace{-8pt}

\paragraph{Implicit field from query-specific features.}
For occupancy prediction at a query point $p$, ConvONet~\cite{Peng2020_CON} extracts convolutional features at $p$ by 
interpolating over a feature field defined for an input shape. 
IF-Net~\cite{Chibane2020} also extracts a feature at $p$ but its associated feature grid encodes multi-scale features
representing both global and local shape information.
Given an input point cloud, Points2Surf~\cite{Points2Surf} samples a local point patch close to the query $p$, as well as
a global point sub-sampling emanating from $p$, and encodes them into two features vectors for implicit decoding.
\rizi{More recently, AIR-Net~\cite{giebenhain2021air}, POCO~\cite{boulch2022poco}, 3DLIG~\cite{zhang20223dilg}, and SA-ConvONet~\cite{tang2021sa}, improve over ConvONet with irregular grids, attention mechanisms, and sign-agnostic optimizations.}

Compared to all these query-specific encodings, ARO captures shape information around the query $p$ in a more contextual and structure-aware
manner. In particular, while Points2Surf provides a ``uni-directional'' observation from $p$, the local observations
by ARO are radially distributed around $p$, providing a more global context. In addition, the inclusion of distance and directional information in ARO reveals a direction 
connection to occupancy prediction, as explained in Section~\ref{sec:aro_rep}. Owing to these distinctive properties of ARO,
ARO-Net consistently produces higher-quality and more faithful reconstructions, as demonstrated in Section~\ref{sec:results}.

%

\vspace{-8pt}

\paragraph{Basis point set.}
Our view-based ARO encoding is related to basis point set (BPS)~\cite{BPS} for point cloud encoding. BPS 
is a fixed-length feature vector computed as the minimal distances from a fixed set of anchors to 
a given point cloud. BPS can serve as a global shape feature, e.g., in place of ARO, for neural implicit reconstruction
from point clouds.

\section{Method}
\label{sec:method}

In this section, we describe and analyze our novel query-specific shape encoding, ARO, in detail.
We first show in Section \ref{sec:aro_rep} that when a given set of anchors can provide a full coverage of a shape, i.e., it observe the entire shape surface, then point occupancies around the shape surface can be fully determined from the corresponding ARO representation.
More generally however, the full coverage condition may not be satisfied or in the case of 3D reconstruction from a noisy and sparse point cloud, the partial observations at the anchors may be inaccurate. In such cases, we train a neural network to predict the point occupancies. This network learns a mapping from the anchored observations to the occupancy value at any spatial query point. We present this network, ARO-Net, in Section \ref{sec:aro_net}. 

\begin{figure*}[!t]
    \centering
    \includegraphics[width=0.98\textwidth]{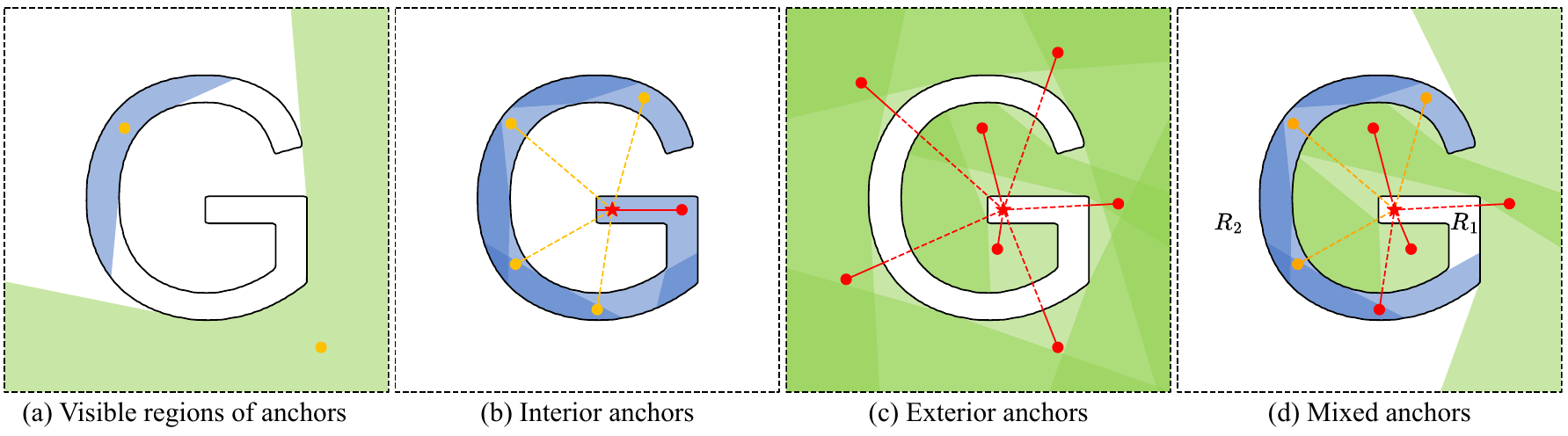}
\caption{Anchor visibility vs. point occupancy. 
The anchors are shown in yellow dots, and the ones that are used to determine the occupancy of the query point (red star) are turn into red.
(a) The visible regions of different anchors relative to the given shape 'G', colored in blue for interior anchors and green for exterior anchors. (b) A set of interior anchors that fully cover the surface, the inside region of the shape is exactly the union of the visible regions of all the anchors. (c) For a set of exterior anchors that fully cover the surface, the region outside the shape is exactly the union of the visible regions of all the anchors. (d) For a set of mixed anchors that together cover the surface, apart from the visible regions, the occupancy of uncovered regions, i.e., $R_1$ and $R_2$, can also be determined. $R_1$ bounded by the surface and the visibility boundary is totally inside the shape, while $R_2$ with dashed virtual boundary is totally outside the shape.}
\label{fig:aro_rep}
\end{figure*}

\subsection{Anchor visibility and point occupancy}
\label{sec:aro_rep}

Given an anchor point $a$, we can cast rays from $a$ in all directions to obtain its visible region relative to a given \rizi{watertight surface} $S$. A ray that does not intersect $S$ can be clipped by the bounding box of the $S$. 
In previous works, such 
visible regions
have been adopted for points inside the shape to assist in shape partition, i.e., the Shape Diameter Function~\cite{SDF}, as well as for points outside the shape for reconstruction~\cite{SSZCO-10}, or to guide relationship optimization between two shapes~\cite{zhao2016relationship}. These works have shown that the 
visible regions
can not only characterize local shape regions, but also provide a global and contextual view of the shape structure.
Figure \ref{fig:aro_rep} (a) shows the 
visible regions of one anchor inside the shape (blue) and the other anchor outside the shape (green). 

Our key observation  to guide the design of our ARO representation is that if we have a set of well-distributed anchors, then combining their 
visible regions together can fully characterize the entire shape. We consider a set of anchors to be well-distributed if any surface point of the shape can be viewed by at least one anchor. 
In the following, we discuss different cases of well-distributed anchors and how they define the shape; see illustrations in Figure \ref{fig:aro_rep}(b-d). 

When the set of well-distributed anchors is all inside the shape as shown in Figure \ref{fig:aro_rep}(b), then the union of all the 
visible regions of the anchors is identical to the interior region of the shape. Thus, a query point, the red star in Figure \ref{fig:aro_rep}(b), is inside the shape if and only if it is {\em covered\/} by at least one radial observation, which we call a \emph{covering anchor}, i.e., the red point in Figure \ref{fig:aro_rep}(b). This is true if the distance between the query point and the covering anchor is smaller than the radial distance of this anchor along the ray direction pointing to the query point. 

When the set of well-distributed anchors is all outside the shape as shown in Figure \ref{fig:aro_rep}(c), the conclusion is similar for the points outside the shape. Thus, the union of all the 
visible regions of the anchors is identical to the region exterior to the shape. Therefore, a query point, i.e., the red star in Figure \ref{fig:aro_rep}(c), is inside the shape if and only if it is {\em not covered by any\/} of the 
visible regions. That is, its distance to each anchor is always larger than the radial distance of this anchor along the ray direction pointing to the query point. Thus, all the anchors together determine the occupancy of the query point. 

When the anchor locations are mixed, with some inside and some outside the shape as shown in Figure \ref{fig:aro_rep}(d), the uncovered spatial regions, $R_1$ and $R_2$, could be either inside or outside the shape. If the region is bounded by the shape boundary and the visible region boundaries for the anchors, such as $R_1$, then the entire region is inside the shape. Locally, this is similar to cases shown in (c) with those red anchors fully covering its boundary. As for $R_2$, the region is partially bounded by the virtual bounding box, i.e., for the image/volume capture, then it is totally outside the shape.

To summarize, if we have a set of anchors that together observe the entire surface of the shape, then 
their visible regions together can fully determine the inside and outside regions of the shape. 
Moreover, one important property is that it converts the inside/outside judgment of a spatial point relative to the shape, to the point's relationship to the radial observations of the anchors, which leads to our definition of query-specific ARO. This property allows to design a network than predicts the occupancy of a spatial point based on \emph{local features} captured by ARO when the anchors are not well-distributed and the shape is only partially observed. 

\subsection{ARO-Net for occupancy prediction}
\label{sec:aro_net}

Given a point cloud $P$, our goal is to reconstruct the corresponding watertight surface $S$ based on our ARO representation \rizi{for point clouds}. We choose to use an implicit representation of the surface, which is essentially a function determining whether a given query point $x$ in space is inside or outside the surface $S$. Unlike previous methods that encode the entire point cloud as a single global latent code~\cite{IM-Net,mescheder2019occupancy}, we utilize the set of anchors to identify local surface regions that are essential to classify occupancy. This allows us to implement a \emph{category-agnostic} method to reconstruct any shape. Figure~\ref{fig:aro_net} shows how query-specific ARO is extracted and utilized by ARO-Net, which outputs the occupancy for any given query point in space.

The input of our method consists of a point cloud $P = \{p_1, \dots, p_n\}$ and an anchor set $A = \{a_1, \dots, a_m\}$ as shown in Figure~\ref{fig:aro_net} (a), where the point cloud are drawn in gray and anchors are drawn with different colors. 
Given a query point $x$ shown as the red dot in Figure~\ref{fig:aro_net} (b), for each anchor $a_i$, we can compute the relative position of $x$ to $a_i$ as $r_i = x - a_i$. 
Next, we create a cone with apex at $a_i$ and $r_i$ as the axis to get the nearest $k$ points $O_i = \{p_i^j\}_{j=1, \dots, k}$ from $P$, which is defined as the radial observation of $a_i$ relative to the query point $x$ and referred as ARO. 
Each ARO $O_i$ is then passed to the PointNet encoder to get the corresponding embedding feature $f_i$. 
The relative location $r_i$ and its norm $\lVert r_i \rVert$ are further concatenated to get a local feature representative of query $x$ relative to anchor $a_i$.
All the local features relative to all the anchors are then passed to an attention module and finally decoded into an occupancy value.

\vspace{-8pt}

\begin{figure}[!t]
    \centering
    \includegraphics[width=0.88\linewidth]{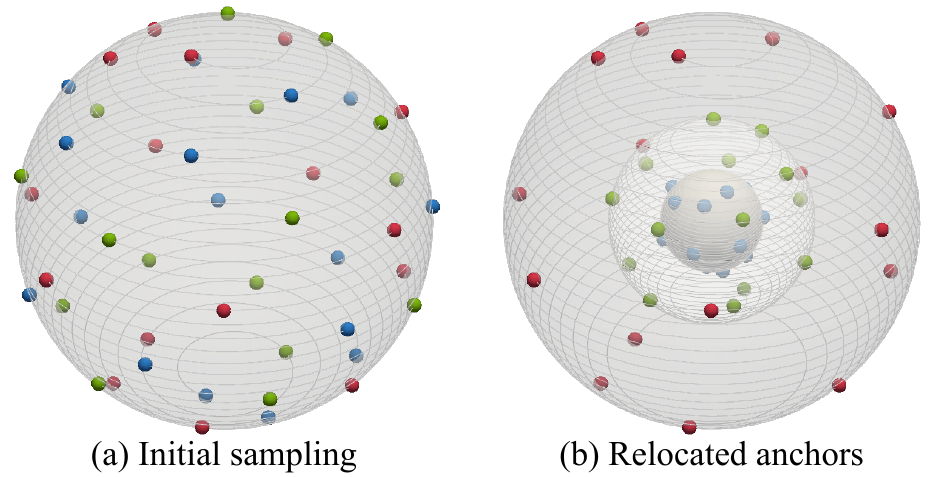}
\caption{Anchor placement.}
\label{fig:anchor}
\end{figure}

\paragraph{Anchors placement.}
A good set of anchor points $A = \{a_1, a_2, \dots, a_m\}$ must be located in different positions in the space bounding the shape, so that they will observe more of the shape's surface with high probability. Hence, to build our set, we sample a set of anchors on three spheres with different radii around the shape. In more detail, assuming that the input point cloud is normalized inside a unit sphere, we first uniformly sample $m$ points on the surface of the unit sphere with radius $r=1/2$ using Fibonacci Sampling~\cite{keinert2015spherical}, then for the points with index $i\%3=1$ we move the point towards the center to position them on the sphere with radius $r = 1/4$, and for the points with index $i\%3=2$ we further move the point to the sphere with radius $r=1/8$, as shown in Figure~\ref{fig:anchor}. This creates a set of anchors distributed in space that can well observe any shape within the sphere. 

\section{Results and evaluation}
\label{sec:results}

\subsection{Experiment setting}

\paragraph{Implementation details.}

\rizi{We employ a binary cross entropy loss for predicted occupancies and an Adam optimizer whose initial learning rate is 3e-4, decaying by 0.5 every 100 epochs.}
The conical angle is set to be $24^{\circ}$ and $k=16$ in all our experiments. The attention module is a Transformer~\cite{vaswani2017attention} encoder consisting of 6 encoding layers and 8 attention heads. The implicit decoder is a fully connected layer which outputs the occupancy value.
We use the sample PointNet architecture as Occ-Net~\cite{mescheder2019occupancy}.
ARO-Net receives 2,048 points as input when training, and can handle arbitrary point numbers when testing.
We tested 3 sets of point numbers $n = \{512, 1024, 2048\}$.

\begin{table}
	\centering
\resizebox{.99\columnwidth}{!}{
	\begin{tabular}{lrrrrr}
		\toprule
        Method & LFD$\downarrow$ & HD$\downarrow$ & CD$\downarrow$   & EMD$\downarrow$  & IOU$\uparrow$ \\
		\midrule
         IM-Net~\cite{IM-Net}  & 3.27  & 11.96  & 57.00 & 11.40 & 3.63\\
         OccNet~\cite{mescheder2019occupancy} & 2.49 & 11.95  & 54.68  & 11.60 & 3.51 \\
         BPS~\cite{BPS} & 2.64 & 5.03 & 11.72  & 2.66 & 6.56 \\
         ConvONet~\cite{Peng2020_CON} & 2.69 & 3.87 & 8.43  & 1.71 & 6.78 \\
         Points2Surf~\cite{Points2Surf} &  1.64 & \underline{2.75}  & 5.69  & 1.25  & \underline{8.36} \\
         UNDC~\cite{NDC} &  \textbf{1.25} & 2.78  & \textbf{4.90}  & \underline{1.17} & 8.21 \\
         ARO-Net  & \underline{1.35} &  \textbf{2.25}  & \underline{5.46} & \textbf{1.12} & \textbf{8.79}
 \\
		\bottomrule
	\end{tabular}
 	}
  	\caption{Quantitative comparisons to state-of-the-art methods on the ABC dataset. 
    Numbers in bold represent first ranking and those underlined in second ranking, in each column. Note that the original version of IM-Net does not take point clouds as input, thus we replace its encoder with OccNet's PointNet encoder.}

    \label{tab:eval_quan_abc}
\end{table}

\begin{figure*}
\centering
  \includegraphics[width=\textwidth]{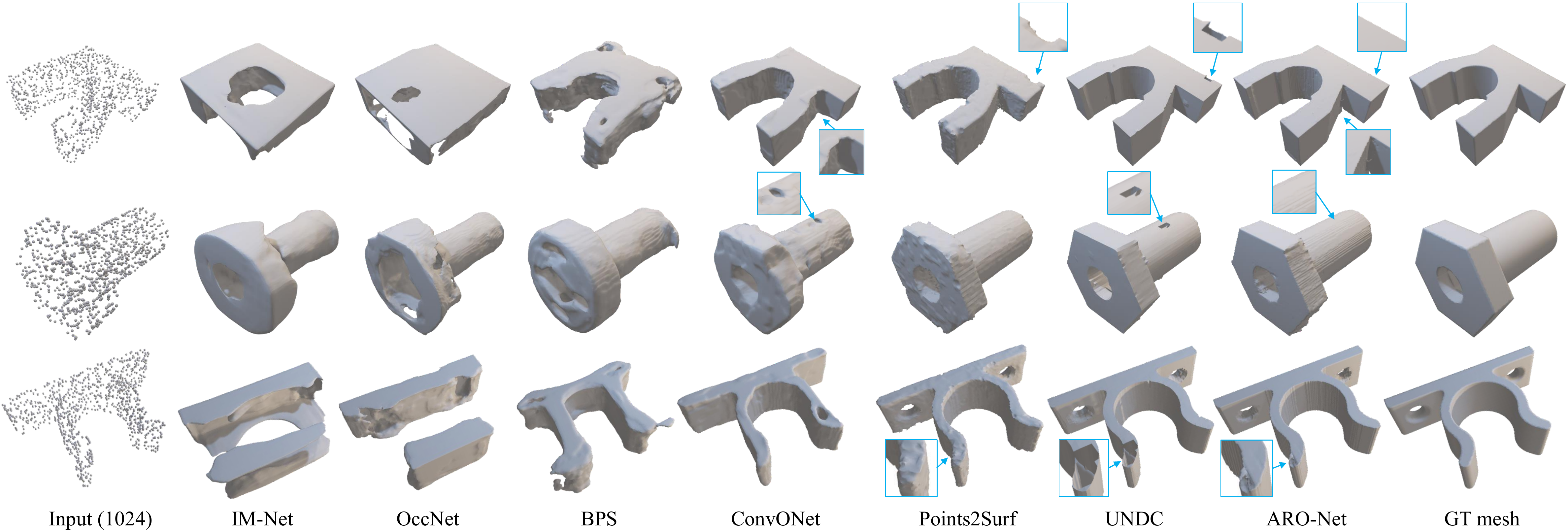}
  \caption{Visual comparisons to state-of-the-art methods on ABC. Zoom-ins highlight reconstruction artifacts to contrast with ARO-Net.}
  \label{fig:eval_qual_abc}
\end{figure*}

\vspace{-8pt}

\paragraph{Dataset.} 
Our experiments are conducted on the ``Big CAD dataset", ABC~\cite{koch2019abc}, and ShapeNet V1~\cite{chang2015ShapeNet}. For ABC, we use the same train/test split as NDC~\cite{NDC}, with 4,011 shapes for training and 982 shapes for testing. For ShapeNet V1, we train ARO-Net using 4K chairs and test 100 shapes per category. Note
that ABC is an especially challenging dataset for 3D representation learning since there is no clear notion of object category (no category labels at all), while the shapes exhibit significant geometry and topology variations.

\vspace{-8pt}

\paragraph{Evaluation metrics.} We adopt Chamfer distance (CD), Hausdoff distance (HD), earth mover's distance (EMD), occupancy intersection over union (IOU) proposed in~\cite{mescheder2019occupancy}, and light filed distance (LFD)~\cite{chen2003visual} as the evaluation metrics for reconstructed meshes.
Note that since the IOU computation method provided by~\cite{mescheder2019occupancy} requires watertight shapes, we report IOU only on the ABC dataset.

\subsection{Comparisons to state-of-the-art methods}


\paragraph{Reconstruction on ABC.}
Quantitative comparisons to state-of-the-art neural 3D reconstruction models on ABC are shown in Table.~\ref{tab:eval_quan_abc}. 
We can see that those methods using only global shape features, i.e., Occ-Net~\cite{mescheder2019occupancy}, IM-Net~\cite{IM-Net}, and  BPS~\cite{BPS}, are consistently outperformed by UNDC, which learns local tessellation, and ConvONet, Points2Surf, and ARO-Net, which all encode local and query-specific features.
Among the last four methods, ARO-Net exhibits a clear advantage in Hausdoff distance and OCC-IOU, and produces the overall best performance by ranking the first in three out of five metrics (IOU, EMD and HD) and ranking the second in one metric (LFD). 
For the remaining CD metric, ARO-Net is still a close runner-up. 

Some visual comparisons are shown in Figure~\ref{fig:eval_qual_abc}.
As shape variations in ABC is large, Occ-Net~\cite{mescheder2019occupancy} and IM-Net~\cite{IM-Net} cannot faithfully reconstruct global shape structures.
With the anchored global features, BPS~\cite{BPS} better captures the overall shapes but fails to reconstruct local details.
ConvONet~\cite{Peng2020_CON} yields smoother local geometry but is unable to reconstruct sharp features.
Points2Surf~\cite{Points2Surf} tends to produce random noises near reconstructed surfaces.
While showing high performance in quantitative evaluations, visual results from UNDC~\cite{NDC} often contain holes with sparse point cloud inputs.
Compared to all those methods, ARO-Net is able to reconstruct both global structure and local detail well, including sharp features as UNDC, and its results
exhibit the least amount of artifacts in terms of holes and noise.

\begin{figure}
\centering
  \includegraphics[width=\columnwidth]{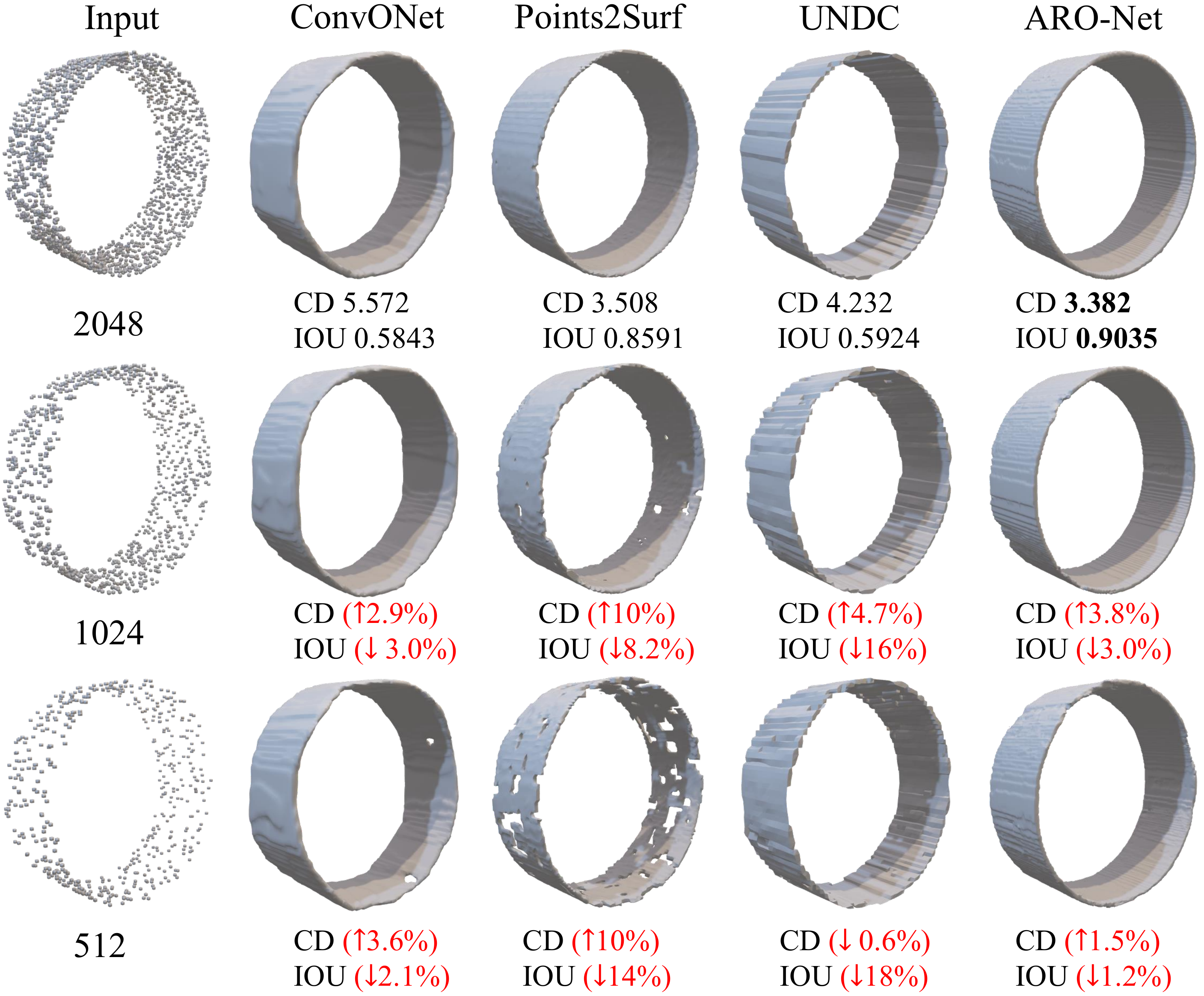}
  \caption{ARO-Net exhibits superior robustness to sparsity of input point clouds compared to its close competitors.}
  \label{fig:sparsity_robustness}
\end{figure}


ARO-Net also exhibits superior robustness to sparsity of input point clouds. Once trained with 2,048-point input, ARO-Net can produce quality results with sparser inputs {\em without re-training\/}. As shown in Figure~\ref{fig:sparsity_robustness}, when the number of points decreases from 2,048 to 512, the reconstruction quality of both UNDC~\cite{NDC} and Points2Surf~\cite{Points2Surf} degrades dramatically, as also reflected in the CD and IOU metrics. Specifically, Points2Surf~\cite{Points2Surf} tends to generate more holes and UNDC~\cite{NDC} produces rougher, oscillatory surfaces. 
ConvONet~\cite{Peng2020_CON} is relatively stable but holes also appear with 512-point inputs. 
By contrast, ARO-Net yields the highest level of robustness both visually and quantitatively.

\begin{table}
	\centering
\resizebox{.85\columnwidth}{!}{
	\begin{tabular}{lrrrr}
		\toprule
		 \multicolumn{5}{c}{Trained on chairs, tested on chairs} \\
		\midrule
        Method & LFD$\downarrow$ & HD$\downarrow$ & CD$\downarrow$  & EMD$\downarrow$   \\
		\midrule
         IM-Net~\cite{IM-Net} & 3.18 & 9.04 & 13.84  & 15.40  \\
         OccNet~\cite{mescheder2019occupancy} & 2.59  & 7.77 & 12.36 & 13.47  \\
         BPS~\cite{BPS}  & 4.31 & 12.28 & 20.51 & 23.30   \\
         ConvONet~\cite{Peng2020_CON} & \underline{2.12} & 6.22 & 11.20 & 12.30  \\
         Points2Surf~\cite{Points2Surf} &  2.51 & 6.46 & 8.60 & \underline{7.34}  \\
         UNDC~\cite{NDC} & 2.19 & \underline{5.60} & \textbf{6.06} & \textbf{6.80}   \\
         ARO-Net & \textbf{1.92} & \textbf{5.33} & \underline{7.14} & 7.70  \\ 
		\midrule
        \multicolumn{5}{c}{Trained on chairs, tested on airplanes} \\
        \midrule
        Method & LFD$\downarrow$ & HD$\downarrow$ & CD$\downarrow$  & EMD$\downarrow$  \\
		\midrule
         IM-Net~\cite{IM-Net} & 14.30 & 19.50 &  44.84 & 41.38 \\
         OccNet~\cite{mescheder2019occupancy} & 12.31  & 18.86 & 39.75 &  38.57  \\
         BPS~\cite{BPS} & 13.73 & 19.48 & 38.10  & 36.61  \\
         ConvONet~\cite{Peng2020_CON}  & 5.69 & 15.98  & 13.50 & 10.75 \\
         Points2Surf~\cite{Points2Surf} & 5.48 & 4.70 & 4.86 & 5.76   \\
         UNDC~\cite{NDC} & \underline{3.62}  & \underline{4.40} & \textbf{3.98}  &  \textbf{3.98}   \\
         ARO-Net & \textbf{3.56} & \textbf{4.32} & \underline{4.61} &  \underline{4.73}   \\
		\bottomrule
	\end{tabular}
	}
 	\caption{Quantitative comparisons to state-of-the-art methods on the ShapeNet V1 dataset. 
 Number in bold means ranking the first and the one with underline means ranking the second in each column. 
 }
    \label{tab:eval_quan_shapenet}
\end{table}

\begin{figure*}
\centering
  \includegraphics[width=\textwidth]{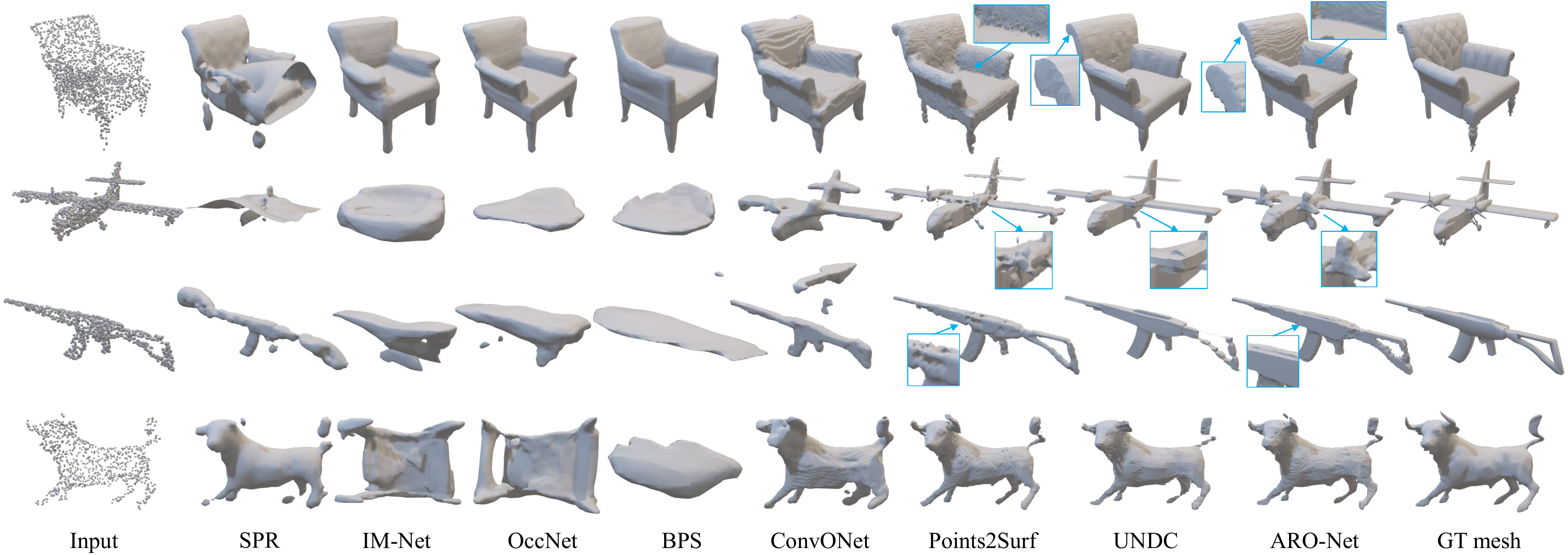}
  \caption{Visual comparisons on ShapeNet V1 and PSB datasets to show reconstruction quality and generalizability of different methods. All methods (except SPR) were trained on chairs in ShapeNet and tested on chairs, airplanes, and rifles in ShapeNet V1 and animals in PSB. Zoom-ins highlight contrasts in details and artifacts in the reconstructions. More results can be found in supplementary material.}
  \label{fig:eval_qual_shapenet}
\end{figure*}

\vspace{-8pt}

\paragraph{Reconstruction on ShapeNet V1.}
For our experiments on the ShapeNet V1 dataset, we train ARO-Net on 4K chairs and test it on other object categories: airplanes and rifle in ShapeNet V1 and animals in PSB~\cite{kalogerakis2010learning}.
Quantitative comparisons are shown in Table~\ref{tab:eval_quan_shapenet}, with some visual results in Figure~\ref{fig:eval_qual_shapenet}.
It is clear that implicit models employing global features do not generalize well. Compared to UNDC~\cite{NDC}, ConvONet~\cite{Peng2020_CON}, and Points2Surf~\cite{Points2Surf}, ARO-Net performs the best overall, quantitatively, especially with the test on airplanes, showing its strong generalizability. 
Visually, ARO-Net produces the most complete, artifact-free results with faithful reconstruction of fine details. In the airplane results in second row, it is worth observing that ARO-Net managed to reconstruct the left-side turbine from extremely sparse points, while other methods all completely missed it.


\vspace{-8pt}

\begin{figure}
\centering
  \includegraphics[width=\columnwidth]{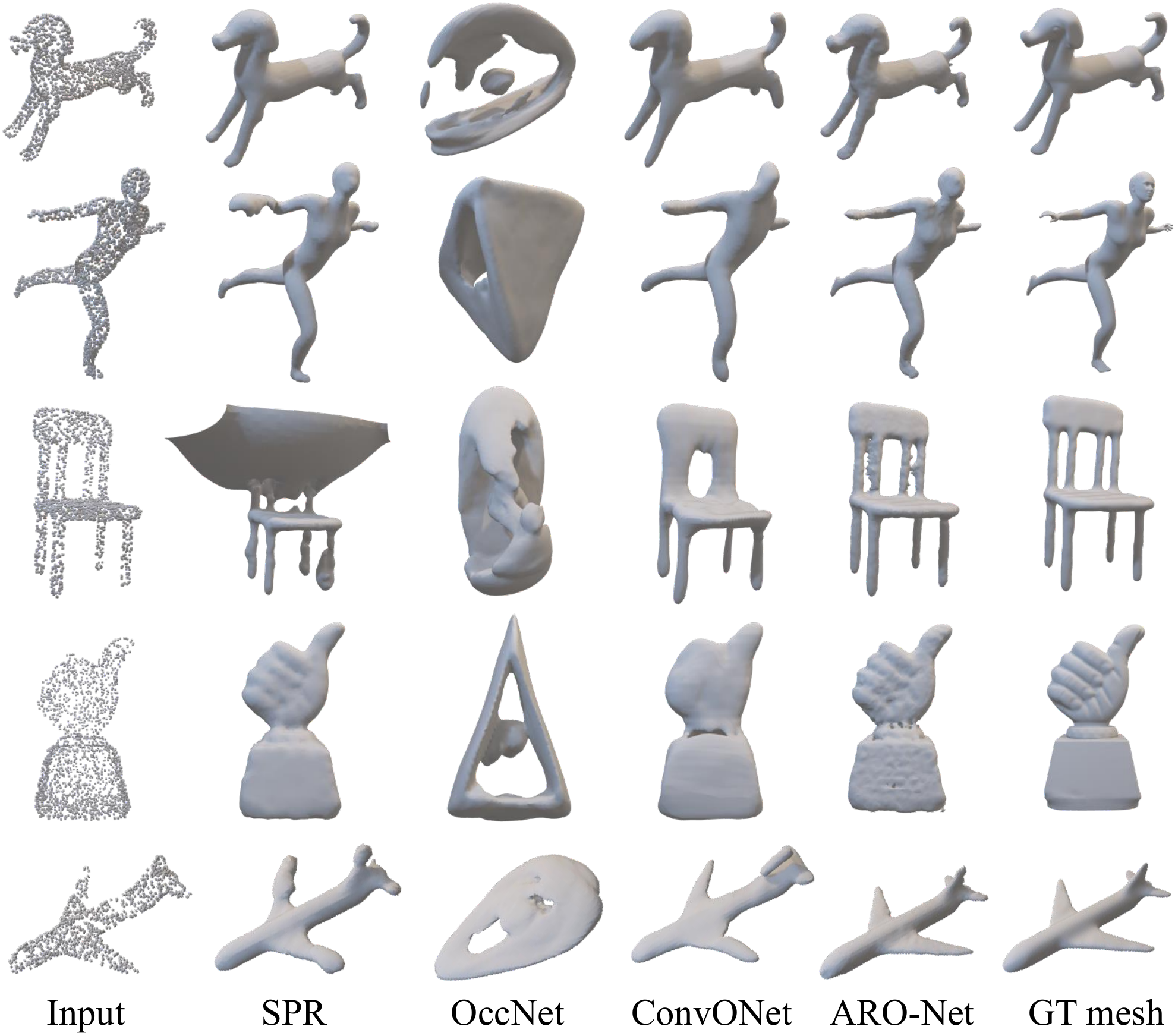}
  \caption{Visual comparison on the results obtained by training only on the ``fertility'' model with rotation and scaling.}
  \label{fig:single_shape_3d_comp}
\end{figure}

\paragraph{One-shape training.}
As a stress test, we train the networks using only one 3D shape, Fertility (see Figure~\ref{fig:teaser}), with rotation and scaling.
From the comparisons in Figure~\ref{fig:single_shape_3d_comp}, we see that SPR suffers from noise and under-sampling (e.g., chair and human hand), as expected. OccNet clearly overfit to the training shape. ConvONet generalizes better but produces overly smoothed results. In contract, ARO-Net reconstructs both global structures and fine details better than the alternatives, with consistence over the shape variety.

\begin{figure}[!t]
    \centering
    \includegraphics[width=0.99\linewidth]{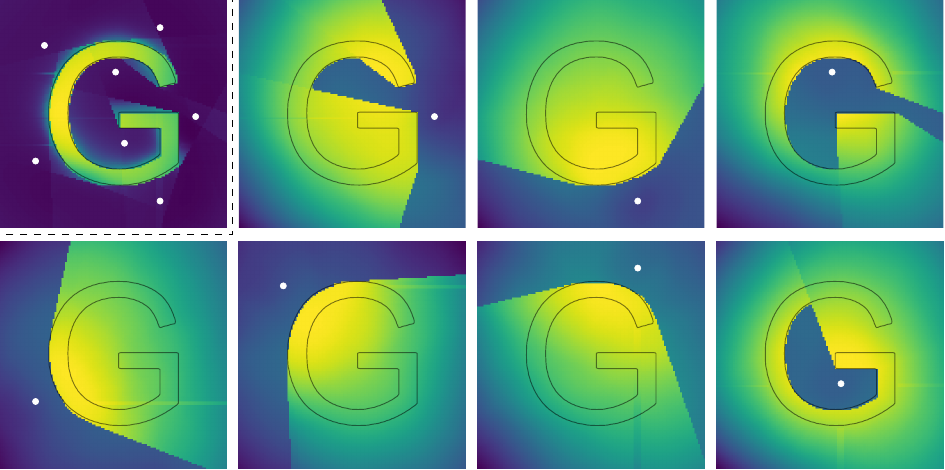}
\caption{Visualization of ARO-Net activation, trained on a single letter ``G'' with seven anchors (top-left). 
The activation for each anchor is closely-related to its visible region.
}
\label{fig:decision}
\end{figure}

\subsection{Ablation study}

\paragraph{Anchor visibility vs. point occupancy.}
To verify whether ARO-Net can learn the close correlation between anchor visibility and point occupancy with the help of a visualization, we implemented a 2D vision of
 ARO-Net with implementation details provided in the supplementary material.
We train ARO-Net using the letter ``G'' with anchors distributed as shown in Figure~\ref{fig:aro_rep}(c). Figure~\ref{fig:decision} visualizes the activation of each anchor by masking out the other anchors' radial observations. The output with all seven anchors considered are shown in the top-left corner. 
The probability of each point being inside the shape is drawn using the blue to yellow colormap, and the anchors are shown in white dots. 
We can see that the decision boundary of each anchor closely follows its visible region, without having that information specifically passed to ARO-Net.
Another noteworthy observation is that for regions invisible to the anchor, points closer to the visibility boundary are assigned higher probability of being inside the shape. This offers evidence that ARO-Net does implicitly learn to predict the point occupancy based on anchor visibility.

\vspace{-8pt}

\paragraph{Anchor placement strategy.}
Other than our layered Fibonacci sampling, we also tried two different sampling strategies: uniform and grid-based sampling.
In uniform sampling, we randomly sample $m$ points in a unit sphere as anchors.
In grid-based sampling, we randomly sample $m$ grid points of a unit cube as anchors.
We found that our layered Fibonacci sampling achieves the best performance.
More detailed comparison can be found in the supplementary material.

\vspace{-8pt}

\paragraph{Anchor count.}
There is a trade-off between reconstruction performance and computational cost when choosing, $m$, the number of anchors. We tested $m = \{24, 48, 96\}$ and report results in the top half of Table~\ref{tab:param_and_ablation}. We found that the performance gain by $m = 96$ is marginal, while the computation cost doubles. With $m = 24$, ARO's performance drops significantly due to lack of sufficient shape coverage. 

\vspace{-8pt}


\paragraph{Decoder architecture.}
The last two rows of Table~\ref{tab:param_and_ablation} compare the current ARO-Net with a version by replacing our attention module and implicit decoder with the much simpler MLPs by IM-Net~\cite{IM-Net}. The similar performance numbers indicate that the superior results by ARO-Net are predominantly owing to ARO rather than the
decoder architecture.

\vspace{-8pt}

\paragraph{Fixed vs.~adaptive anchors.}
\rizi{When computing the radial observations, we perform top-$k$ selection (pooling) on the cosine and Euclidean distances between the anchors and input points. Therefore, our network loss function is differentiable with respect to anchor coordinates: the gradients are passed through the indices that form the top $k$ values, and then passed to anchor coordinates.}
While this does offer the possibility of fine-tuning the anchor positions during training and testing, we have found that optimizing anchor positions along with the network parameters can make the training process unstable and lead to worse results. Overall, we believe that with the anchors fixed, it is easier for the network to learn the mapping from ARO to point occupancy.

During inference, it is possible to adjust the anchor locations to obtain a larger coverage of the shape surface to improve reconstruction. However, we found the improvements to be marginal; see supplementary material.

\begin{table}
	\centering
	\resizebox{.85\columnwidth}{!}{
	\begin{tabular}{lrrrrrr}
		\toprule
		Setting & LFD$\downarrow$ & HD$\downarrow$ & CD$\downarrow$  & EMD$\downarrow$  & IOU$\uparrow$ \\
		\midrule
		$m=24$  & 1.48 & 2.33  &  5.94 & 1.28 & 8.66 \\
         $m=96$ & 1.30 & 2.16  &   5.31 & 1.03 & 8.92 \\
         MLP & 1.44 &  2.30 &  5.90  & 1.14 &  8.75 \\
         Default & 1.35 &  2.25  & 5.46 & 1.12 & 8.79 \\
		\bottomrule
	\end{tabular}
	}
 \caption{Ablation studies on ABC to evaluate our choices for
the anchor count ($m$) and decoder architecture. Default: $m=48$ with attention modules in implicit decoder.}
    \label{tab:param_and_ablation}
\end{table}	

\vspace{-8pt}

\paragraph{Distance and direction information in ARO.}
At last, we have confirmation that the $r$ and $\lVert r \rVert$ play an important role as input to ARO-net. The reconstruction performance by the network drops significantly when we remove such information, with only the PointNet features. Again, more details can be found in the supplementary material.


\section{Conclusion, limitation, and future work}
\label{sec:future}


We introduce a novel neural implicit representation for 3D shapes that is conditioned on 
Anchored Radial Observations (ARO). We use ARO to train our reconstruction network, ARO-Net,
that outputs an occupancy at any query point in space. ARO is query-specific, defined 
by a set of local descriptors from the perspective of a set of fixed anchors to provide 
a global context. ARO-Net is category-agnostic and provides superior 
reconstruction results to alternative methods. It also has strong generalization capabilities 
and can even be used to learn from a single shape.

Some current limitations and future works are as follows. The quality and efficiency of the resulting representation are dependant on the number and placement of the anchors. We tested a number of configurations and settled on a fixed set of $48$ anchors via Fibonacci sampling. However, further research into anchor selection, even learning, is needed. In addition, the ARO-Net representation still requires storing the original input point set and therefore is less memory efficient than pure network-based representations. Lastly, both training and inference of ARO-Net are slower than prior implicit models based on simpler and less contextual shape encodings (e.g.\ ConvONet). This can be alleviated by adaptive sampling of the anchors and using more efficient spatial search data structures to find closest points.


{\small
\bibliographystyle{ieee_fullname}
\bibliography{aro}
}

\clearpage

\newpage%
\twocolumn[\appendixhead]
\appendix
\renewcommand{\thesection}{\Alph{section}}

\section{Anchor Placement}

\begin{figure}
\centering
  \includegraphics[width=0.99\columnwidth]{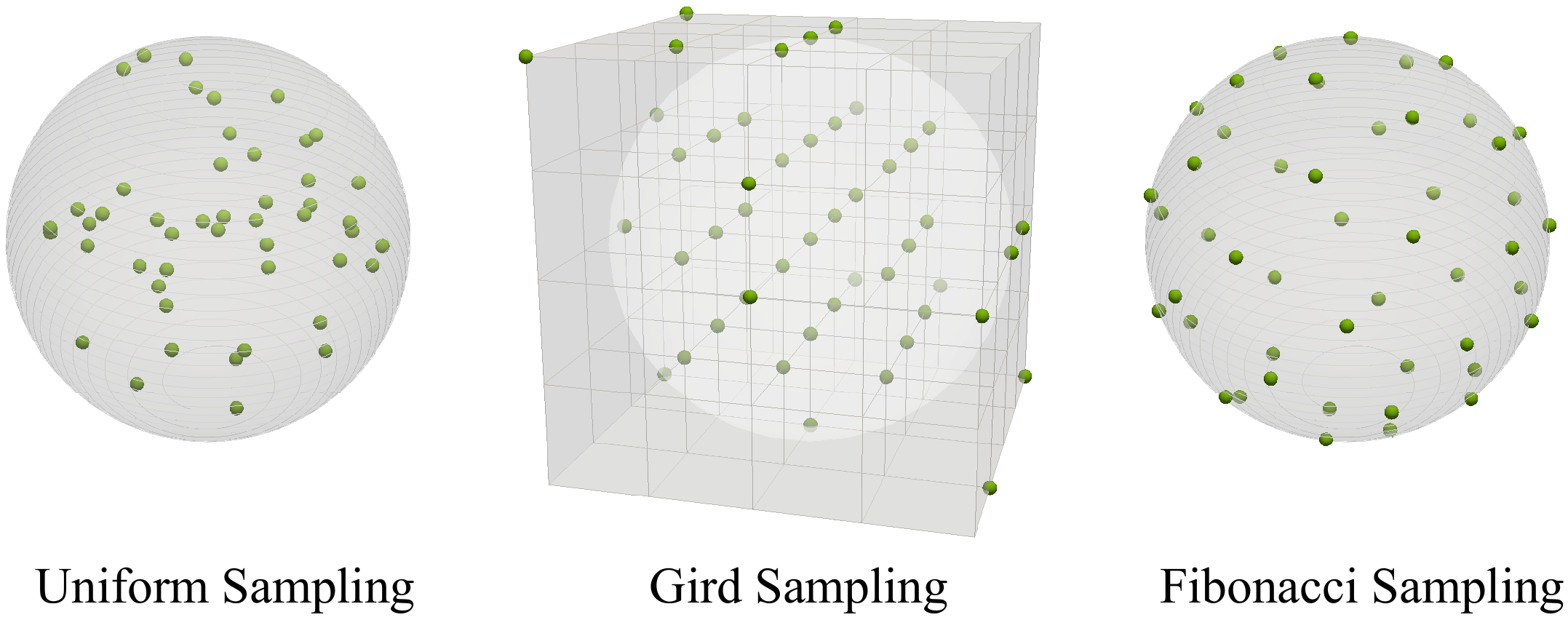}
  \caption{Visualization of different anchor placement strategies.}
  \label{fig:anchor_sampling_vis}
\end{figure}

\begin{figure}
\centering
  \includegraphics[width=\columnwidth]{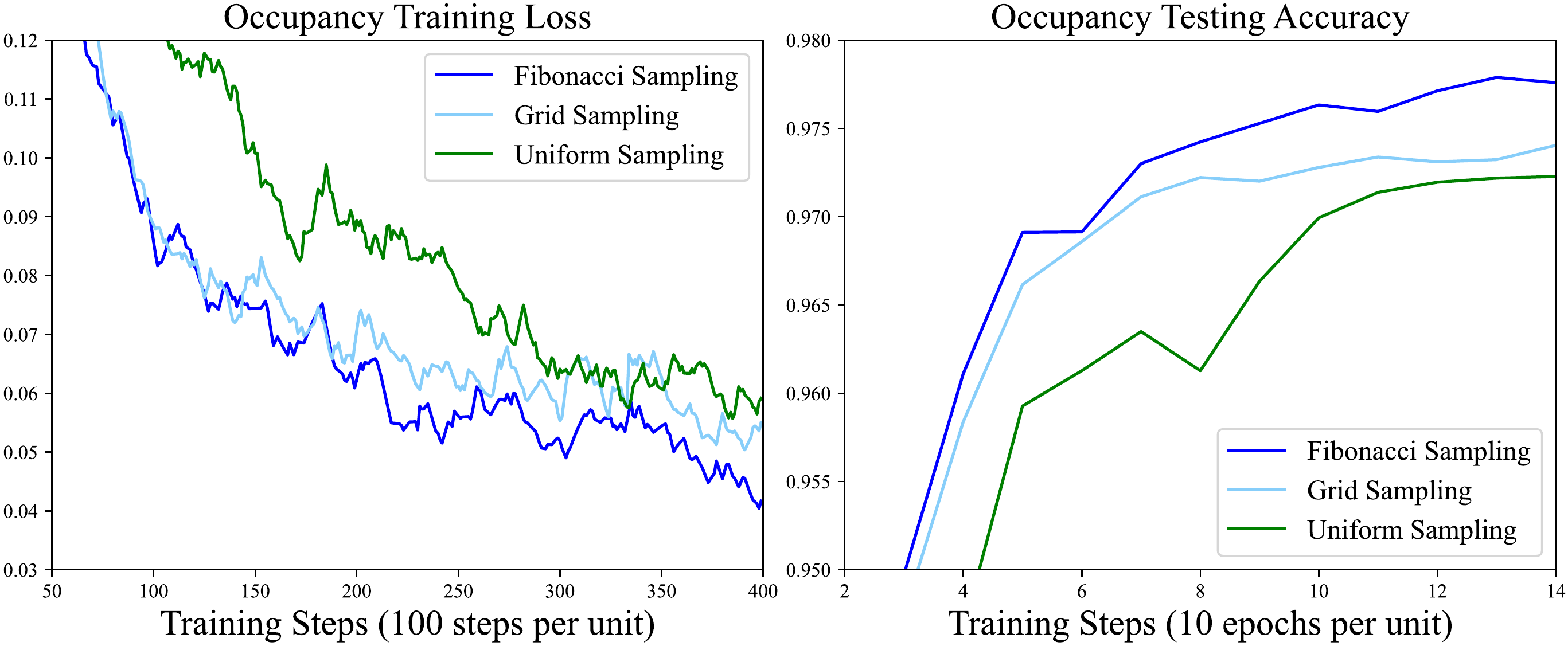}
  \caption{How different anchor placement strategies affect network performance, where anchor count $m = 48$ for all settings.}
  \label{fig:anchor_ablation}
\end{figure}

\begin{figure}
\centering
  \includegraphics[width=0.8\columnwidth]{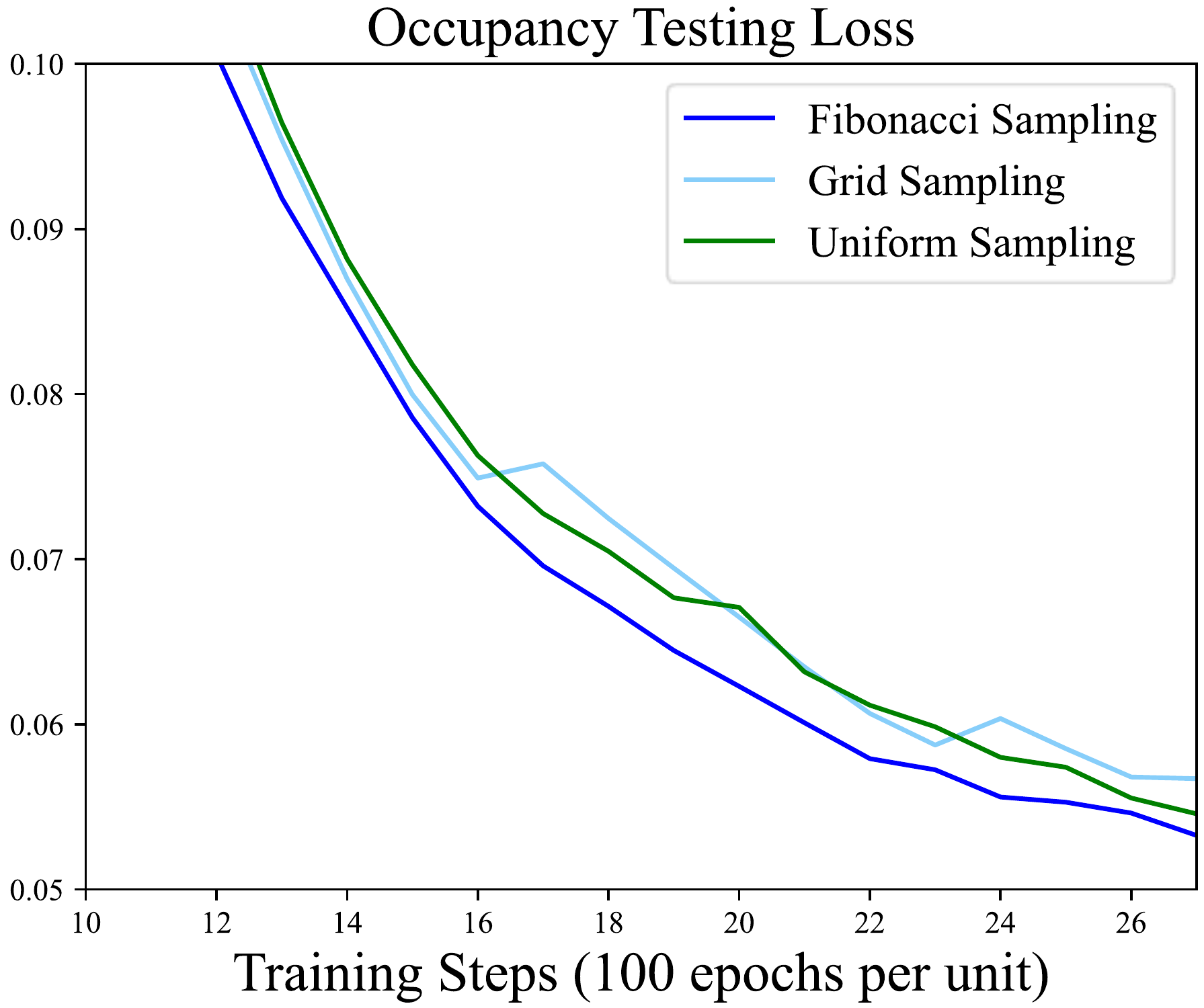}
  \caption{The testing loss for different strategies of anchor placement. The loss is calculated by the cross-entropy between predicted probability and GT occupancy.
}
  \label{fig:loss_and_acc_curves}
\end{figure}

In Figure~\ref{fig:anchor_sampling_vis}, we demonstrate how anchors are placed by uniform, grid and Fibonacci samplings. 
In grid sampling, we first choose all grid points $\{(x, y, z)| x,y,z \in \{-0.5, -0.25, 0, 0.25, 0.5\}\}$ inside the unit sphere and then randomly select from remaining grid points to make a total count of 48.
Table~\ref{tab:anchor_placement} shows how different anchor placement strategies affect the construction performance on ABC, where Fibonacci sampling is slightly better than grid sampling, and significantly better than uniform sampling.
In the early stage of training, Fibonacci sampling exhibits a clear advantage over other two samplings, as shown in Figure~\ref{fig:anchor_ablation}.
In Figure~\ref{fig:loss_and_acc_curves}, we show the curves in the later stage (from Epoch 100 to 260).
We find that grid and uniform sampling narrow their performance gap between Fibonacci sampling in the later stage.
However, as shown in the testing loss curve, gird sampling suffers from turbulence of the loss values.
In terms of the stability of training process and the overall performance, Fibonacci sampling is superior to other two sampling strategies. We leave more comprehensive investigations on anchor placement as our future work.

\section{Adaptive Anchors}

As suggested in Table~\ref{tab:adaptive_anchors}, optimizing the anchor positions in the training stage (the first row) cannot improve the performance of ARO-Net on ABC.
In the testing phase, we try to adjust the anchor positions for each testing case. Specifically, we fix the network parameters, utilize input point clouds as ground-truth occupancies, and then adjust the anchor positions using Gradient Descent algorithm.
Table~\ref{tab:adaptive_anchors} demonstrates that fine-tuning the anchors for 100 iterations can bring very marginal improvements; 200 iterations will degrade the performance because of over-fitting.

\begin{table}
	\centering
	\resizebox{\columnwidth}{!}{
	\begin{tabular}{lrrrrrr}
		\toprule
		Setting & LFD$\downarrow$ & HD$\downarrow$ & CD$\downarrow$  & EMD$\downarrow$  & IOU$\uparrow$ \\
		\midrule
         Uniform & 1.52 & 2.32  & 6.29   & 1.18 & 8.67 \\
         Gird & 1.42 & 2.20  &  5.63  & 1.11 &  8.76 \\
         Fibonacci & 1.35 &  2.25  & 5.46 & 1.12 & 8.79 \\
		\bottomrule
	\end{tabular}
	}
 \caption{How different anchor placement strategies affect the reconstruction performance.}
    \label{tab:anchor_placement}
\end{table}

\begin{table}
	\centering
	\resizebox{.99\columnwidth}{!}{
	\begin{tabular}{lrrrrrr}
		\toprule
		Adjusting Stage & LFD$\downarrow$ & HD$\downarrow$ & CD$\downarrow$  & EMD$\downarrow$  & IOU$\uparrow$ \\
		\midrule
         Training & 1.60 & 2.39  &  5.98  & 1.333 &  8.63 \\
         Testing (100) & 1.58 & 2.24  & 5.44 & 1.098  & 8.78  \\
         Testing (200) & 1.80  &  2.20 &  5.75  & 1.314 &  8.74 \\
         Default (Fixed) & 1.35 &  2.25  & 5.46 & 1.121 & 8.79 \\
		\bottomrule
	\end{tabular}
	}
 \caption{Adjusting anchor positions in the training or testing stage. The numbers behind ``testing'' denote the iteration numbers of adjusting anchor positions.}
    \label{tab:adaptive_anchors}
\end{table}

\section{Distance and Direction Information}

\begin{table}
	\centering
	\resizebox{.99\columnwidth}{!}{
	\begin{tabular}{lrrrrrr}
		\toprule
		Setting & LFD$\downarrow$ & HD$\downarrow$ & CD$\downarrow$  & EMD$\downarrow$  & IOU$\uparrow$ \\
		\midrule
         w/o $r$ and $\lVert r \rVert$ & 1.38 & 2.38  & 5.94  & 1.22  & 8.62 \\

         Full model & 1.35 &  2.25  & 5.46 & 1.12 & 8.79 \\
		\bottomrule
	\end{tabular}
	}
 \caption{Removing the distance ($\lVert r \rVert$) and direction information ($r$) in ARO-Net.}
    \label{tab:remove_dist_and_dir}
\end{table}	

As shown in Table~\ref{tab:remove_dist_and_dir}, the reconstruction performance of ARO-Net drops significantly when we remove $r$ and $\lVert r \rVert$.
Since the prediction of ARO-Net is based on engaging $r$ and $\lVert r \rVert$ with the radial observations, missing $r$ and $\lVert r \rVert$ as network input would make the prediction more difficult.

\begin{figure*}
\centering
  \includegraphics[width=0.95\textwidth]{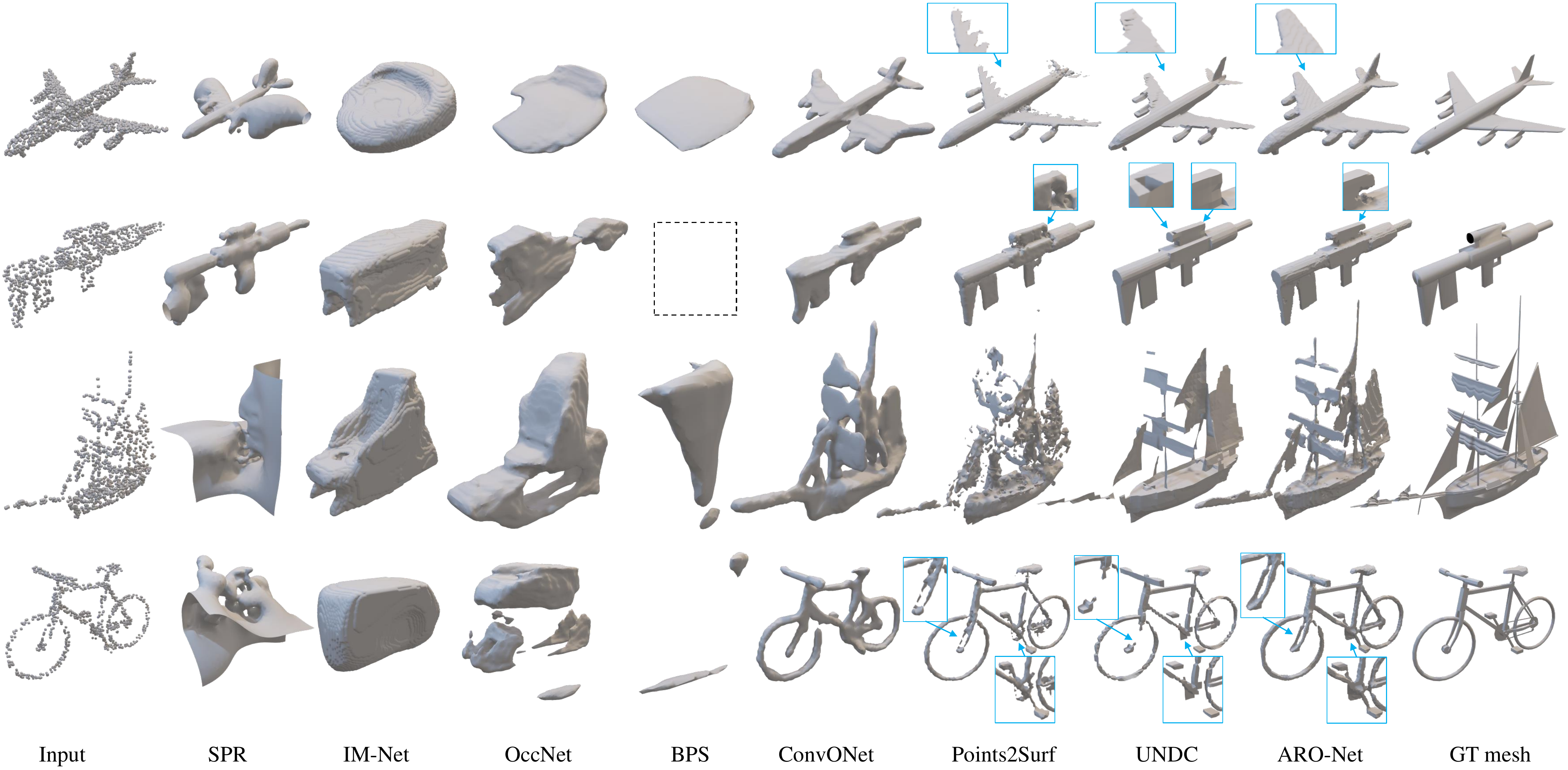}
  \caption{More visual comparisons to state-of-the-art methods on ShapeNet. All methods (except SPR) were trained on chairs in ShapeNet and tested on other categories with zoom-ins highlighting reconstruction artifacts. BPS generated an empty result for the second case.}
  \label{fig:eval_qual_shapenet_sm}
\end{figure*}

\section{2D Experiments}
In 2D experiments, we replace the PointNet feature $f_{i}$ with the hit distance of $r_{i}$ (a ray from anchor $a_{i}$ to query point) on a given 2D shape $S$.
Specifically, the hit distance (denoted as $d_{i}$) is calculated from $a_{i}$ to the hit point on the boundary pixels of $S$.
Note that it is still non-trivial to reconstruct the 2D shape under this setting because (1) the occupancy of non-boundary points remains unknown and (2) the anchors could observe only partial boundaries because of their count and placement.
We send $r$, $\lVert r \rVert$, and $d$ into the network to predict the occupancy.
The attention module consists of 3 encoding
layers and 4 attention heads. When reconstructing the images of 2D shapes, we utilize the predicted occupancy values to construct the pixel values.\par

\section{More Comparisons}
We compare ARO-Net with a more recent work POCO~\cite{boulch2022poco} using LFD ($\downarrow$) in Table~\ref{tab:aronet_vs_poco}, where ARO-Net has a big advantage in both ABC and ShapeNet Chairs/Airplanes.

\begin{table}
\centering
\begin{tabular}{cccc}
    \toprule
    Method & ABC & Chairs & Airplanes \\
    \midrule
     POCO/\textbf{ARO-Net}  & 1.71/\textbf{1.35} & 2.82/\textbf{1.92}  & 5.l1/\textbf{3.56}  \\
    \bottomrule

\end{tabular}
 \caption{Comparison between ARO-Net and POCO.}
    \label{tab:aronet_vs_poco}
\end{table}

\begin{figure*}
\centering
  \includegraphics[width=0.94\textwidth]{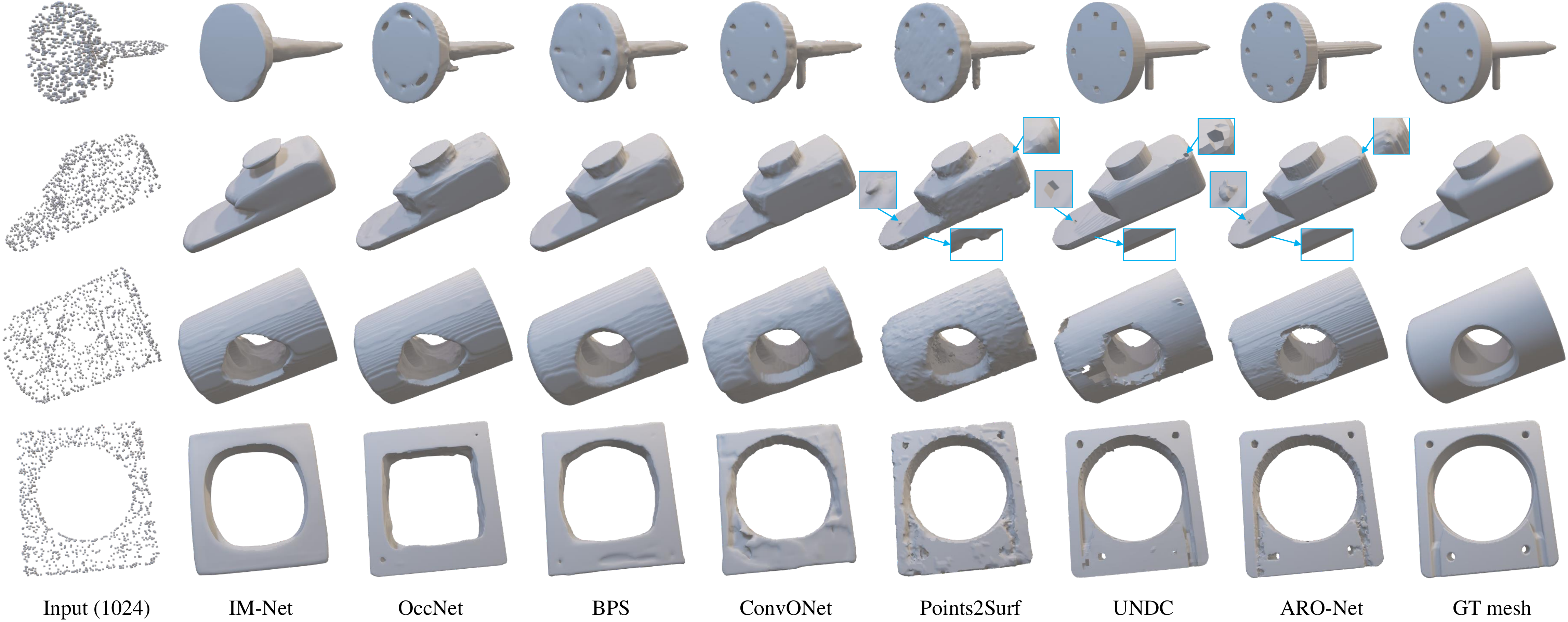}
  \caption{More visual comparisons to state-of-the-art methods on ABC with zoom-ins highlighting reconstruction artifacts.}
  \label{fig:eval_qual_abc_sm}
\end{figure*}

\begin{figure*}
\centering
  \includegraphics[width= 0.94\textwidth]{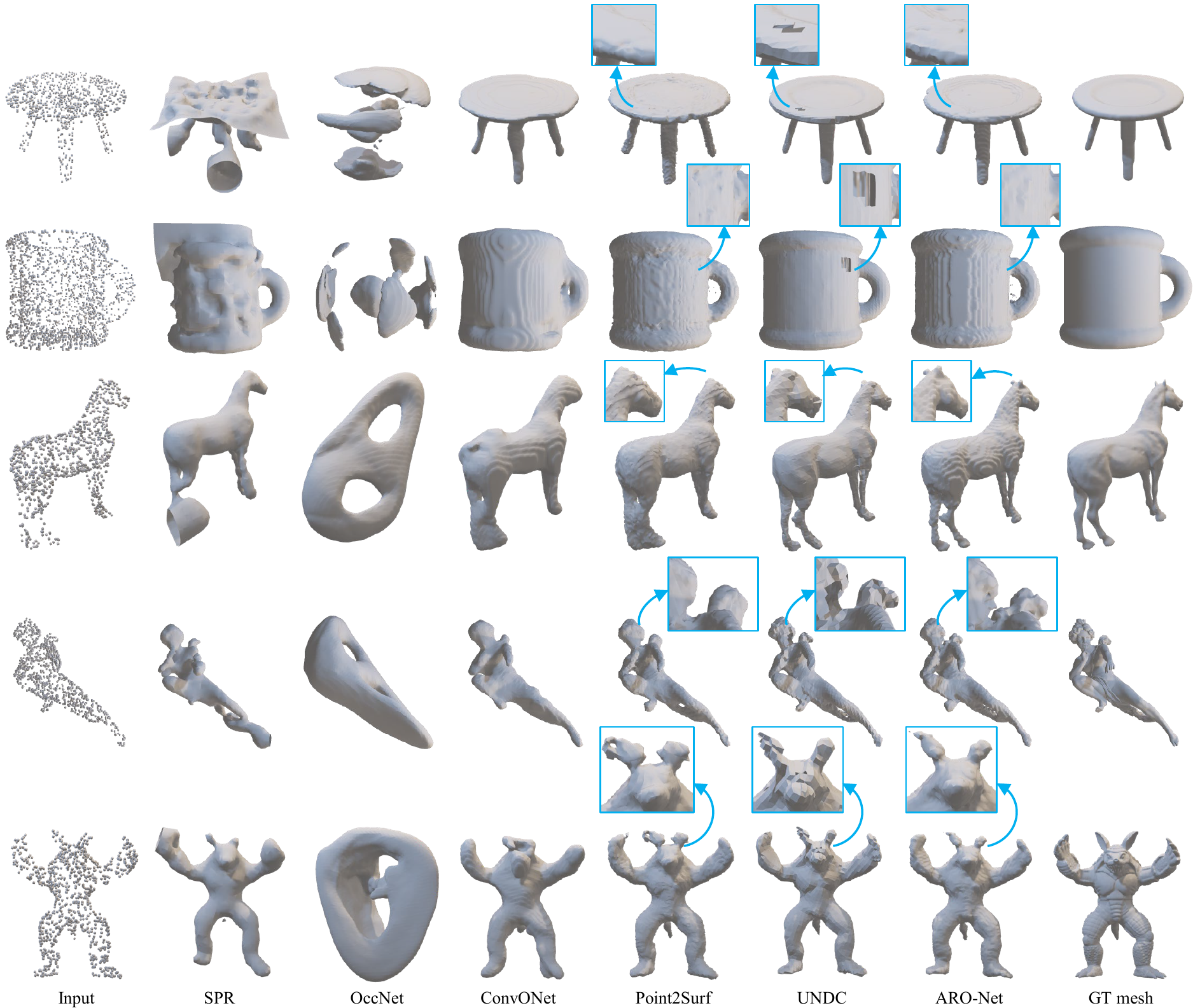}
  \caption{More visual comparisons on the results obtained by training only on the ``fertility'' model with rotation and scaling.}
  \label{fig:eval_qual_oneshape}
\end{figure*}

In Figure~\ref{fig:eval_qual_shapenet_sm} and Figure~\ref{fig:eval_qual_abc_sm}, we provide more visual comparisons on ShapeNet and ABC dataset. In Figure~\ref{fig:eval_qual_oneshape}, we provide more visual comparisons on one-shape training, and adding Points2Surf~\cite{Points2Surf} and UNDC~\cite{NDC} into comparisons.
ARO-Net performs significantly better than others in extremely sparse point clouds (such as the ship and bicycle in Figure~\ref{fig:eval_qual_shapenet}).
ARO-Net also shows great ability in preserving local details such as the holes in 1st and 4th examples in Figure~\ref{fig:eval_qual_abc} and the ears of horse in Figure~\ref{fig:eval_qual_oneshape}.
Compared to UNDC, ARO-Net can avoid generating undesired holes on the surfaces.
Compared to Points2Surf and ConvONet, ARO-Net can reconstruct better details, smoother surfaces and sharper features.
Among all of these comparisons across a variety of shapes, ARO-Net achieves the best overall performance.

\end{document}